\documentclass[letterpaper, 10 pt, journal, twoside]{ieeetran}
\usepackage{tabularx}
\usepackage{amsmath}
\usepackage{hyperref}
\usepackage{mathtools}
\usepackage{subcaption}
\usepackage{annotate-equations}
\usepackage{cleveref}
\usepackage{siunitx}
\usepackage{url}
\usepackage{fontawesome5}
\usepackage{booktabs}
\usepackage{multirow}
\usepackage{xcolor, colortbl}
\usepackage{hyperref} %
\usepackage{cuted}
\usepackage{capt-of}
\usepackage{censor}

\definecolor{lightgray}{gray}{0.92}

\definecolor{redhighlight}{RGB}{255, 0, 0}
\definecolor{greenhighlight}{RGB}{0, 255, 0}
\definecolor{bluehighlight}{RGB}{0, 0, 255}

\definecolor{mpc_color}{RGB}{193, 48, 55}
\definecolor{mpg_color}{RGB}{32, 112, 186}
\definecolor{lmpg_color}{RGB}{219, 194, 35}
\definecolor{lmpggen_color}{RGB}{176, 156, 26}
\definecolor{mpccp_color}{RGB}{255, 140, 0}

\definecolor{redbg}{rgb}{255,0,0}
\colorlet{redbg}{redbg!75}
\definecolor{bluebg}{rgb}{0,0,255}
\colorlet{bluebg}{bluebg!75}

\newcommand{\BLUE}[1]{\textcolor{black}{#1}}

\usepackage{graphics}           
\usepackage{times}              
\usepackage{amsmath}            
\usepackage{amssymb}            
\usepackage{graphicx}
\usepackage{algorithm}
\usepackage[noend]{algpseudocode}
\usepackage{booktabs}
\usepackage{color}
\definecolor{instructioncolor}{rgb}{.5,.5,.5}

\usepackage[font=small]{caption}

\def\figref#1{Fig.~\ref{#1}}
\def\tabref#1{Tab.~\ref{#1}}
\def\eqref#1{Eq.~(\ref{#1})}

\makeatletter
\usepackage{xspace}
\DeclareRobustCommand\onedot{\futurelet\@let@token\@onedot}
\def\@onedot{\ifx\@let@token.\else.\null\fi\xspace}

\makeatother

\usepackage{array}
\newcolumntype{L}[1]{>{\raggedright\let\newline\\\arraybackslash\hspace{0pt}}m{#1}}
\newcolumntype{C}[1]{>{\centering\let\newline\\\arraybackslash\hspace{0pt}}m{#1}}
\newcolumntype{R}[1]{>{\raggedleft\let\newline\\\arraybackslash\hspace{0pt}}m{#1}}

\title{Strategizing at Speed: A Learned Model Predictive Game for Multi-Agent Drone Racing}

\author{Andrei-Carlo Papuc, Lasse Peters, Sihao Sun, Laura Ferranti, and Javier Alonso-Mora %
\thanks{Manuscript received: February, 5, 2026; Revised April, 29, 2026; Accepted May, 25, 2026. This paper was recommended for publication by Editor Wei Pan upon evaluation of the Associate Editor and Reviewers’ comments.
This work was supported by the Office of Naval Research Global under Grant N62909-25-12027 (Project SECURE) and by the Dutch Research Council (NWO) under Grant no. 20256, (Project Accurate Aerial Manipulation).} %
\thanks{All authors are with the Department of Cognitive Robotics, Delft University of Technology, 2628 CD Delft, The Netherlands. (Corresponding author: Andrei-Carlo Papuc~{\tt\footnotesize a.c.papuc@tudelft.nl})}
\thanks{Digital Object Identifier (DOI): see top of this page.}
}

\usepackage{titlesec}
\titlespacing*{\subsection}{0pt}{1.5ex}{1.5ex}

\begin{document}
\maketitle

\begin{abstract}
Autonomous drone racing pushes the boundaries of high-speed motion planning and multi-agent strategic decision-making.
Success in this domain requires drones not only to navigate at their limits but also to anticipate and counteract competitors' actions.
In this paper, we study a fundamental question that arises in this domain: how deeply should an agent strategize before taking an action?
To this end, we compare two planning paradigms: the Model Predictive Game (MPG), which finds interaction-aware strategies at the expense of longer computation times, and contouring Model Predictive Control (MPC), which computes strategies rapidly but does not reason about interactions.
We perform extensive experiments to study this trade-off, revealing that MPG outperforms MPC at moderate velocities but loses its advantage at higher speeds due to latency.
To address this shortcoming, we propose a Learned Model Predictive Game (LMPG) approach that amortizes model predictive gameplay to reduce latency.
In both simulation and hardware experiments, we benchmark our approach against MPG and MPC in head-to-head races, finding that LMPG outperforms both baselines. ~\href{https://github.com/andrejcarlo/ral26_strategizing_at_speed/}
     {\textcolor{blue!60!black}{\faGithub\enspace{Code}}} ~\href{https://youtu.be/JalS5aQ7LWA}
     {\textcolor{blue!60!black}{\faYoutube\enspace{Video}}}
\end{abstract}


\section{Introduction}
\label{sec:intro}

\IEEEPARstart{A}{utonomous} racing serves as a rigorous testbed for intelligent systems operating at the physical limits of handling. To compete at or above human performance levels, these systems must navigate complex 3D environments at high speeds while making real-time strategic decisions against adversarial opponents.
Advancements in the fields of planning and control are necessary for this, as demonstrated in previous research focused on time-optimal flight \cite{song2021autonomous, kaufmann2023champion, ferede2024end}.

\begin{figure}[t]
    \centering
    \includegraphics[width=1\linewidth]{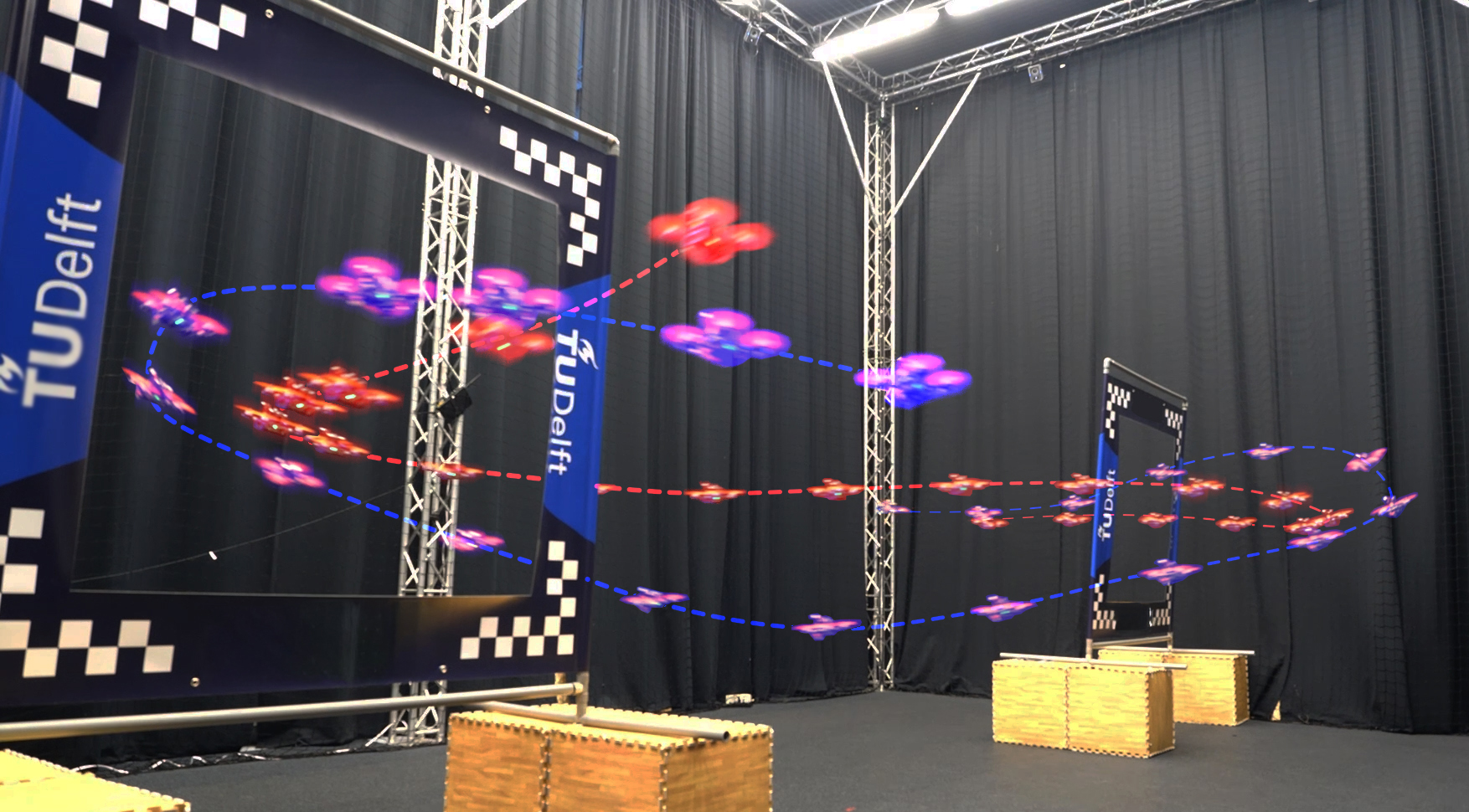}
    \caption{Chronophotography of a real-world autonomous overtaking maneuver on a lemniscate track. The interaction-aware blue drone employs a game-theoretic planner to overtake the red drone through the gate. The red drone utilizes contouring MPC with a constant velocity prediction model, treating the opponent as a dynamic obstacle.}
    \label{fig:motivation}
    \vspace{-1.0em}
\end{figure}

While much of the recent research in autonomous racing has focused on optimizing single-agent performance \cite{hanoverAutonomousDroneRacing2024}, such as minimizing lap times, real-world racing scenarios often involve multiple competitors, each with their own strategies and goals.
This creates a dynamic, multi-agent environment where decision-making is influenced by the actions of other participants, see~\figref{fig:motivation} for a race.

Conventional approaches like Model Predictive Control (MPC) typically rely on a \textit{predict-then-plan} scheme \cite{rowoldEfficientSpatiotemporalGraph2022, jank2023hierarchical, linigerOptimizationbasedAutonomousRacing2015}, treating opponents as non-reactive dynamic obstacles.
These methods ignore the fundamental interdependence between agents, instead predicting a fixed opponent trajectory independent of the ego vehicle's actions.
This can lead to suboptimal or overly conservative behavior, as the planner fails to anticipate defensive or aggressive counter-maneuvers.

Model predictive game (MPG) planners address this limitation by explicitly modeling coupled dynamics and intents~\cite{spicaRealTimeGameTheoretic2018, lecleachALGAMESFastAugmented2022, zhuSequentialQuadraticProgramming2024}.
However, their strategic reasoning comes at a high computational cost. 
Since the state evolves while the drone is~``thinking,'' computation delays can cause strategies to be outdated by the time they are applied.
We find that this added computational overhead often negates the competitive advantage of MPG in real-world experiments. %

Motivated by this observation, this work examines the trade-off between extensive planning to find high-fidelity strategies and expedited decision-making that ignores future interdependence of actions.
The main contribution of this paper is the Learned Model Predictive Game (LMPG), a method that efficiently solves dynamic games online by leveraging an amortized game formulation to accelerate computation. Building on the approach of~\cite{petersLearningMixedStrategies2022}, we introduce a specialized training procedure tailored to the high-speed dynamics of racing applications.
In both simulation and hardware experiments, we benchmark our approach against MPG and MPC variants in head-to-head races, finding that LMPG outperforms both baselines.

\section{Related Work}
\label{sec:related}

Methods for model-based multi-agent autonomous racing generally fall into two categories: optimization-based planners that treat opponents as obstacles, and game-theoretic approaches that explicitly model adversarial intent. %

\smallskip\noindent\textbf{Model predictive control (MPC).}
MPC approaches typically view opponents as static or dynamic obstacles with fixed trajectories obtained from a prediction model \cite{hanoverAutonomousDroneRacing2024}.
To handle the computational demands of high-speed racing, recent works have introduced hierarchical frameworks.
For instance, \cite{linigerOptimizationbasedAutonomousRacing2015} utilizes a high-level planner for feasible trajectory generation followed by a non-linear MPC for tracking.
Similarly, methods incorporating spatiotemporal graph search \cite{rowoldEfficientSpatiotemporalGraph2022} or hybrid learning-based local planners \cite{evansLearningSubsystemLocal2021} have been developed to improve obstacle avoidance and lap times.
However, these methods inherently decouple the ego-vehicle’s planning from the opponent’s decision-making, often leading to conservative behavior that fails to execute strategic maneuvers, such as blocking or aggressive overtaking.

\smallskip\noindent\textbf{Model predictive games (MPG).}
To address the lack of interactive reasoning, MPG planners model the race as a dynamic game, solving for Nash or Stackelberg equilibria at every planner invocation.
Through this formulation, these methods account for the fact that agents' future actions are interdependent.
In some cases, noncooperative games can be formulated as \emph{potential} games~\cite{jiaRAPIDAutonomousMultiAgent2023}, facilitating solution via off-the-shelf (single-player) optimization techniques.
In general, however, noncooperative games cannot be reduced to a single optimization problem, requiring specialized solvers for equilibrium search.
Iterated Best Response (IBR) methods \cite{spicaRealTimeGameTheoretic2018, wangMultiagentSensitivityEnhanced2020, williamsAutonomousRacingAutoRally2017} approximate Nash equilibria by alternating optimization between agents. %
By contrast, coupled approaches, such as ALGAMES~\cite{lecleachALGAMESFastAugmented2022} and sequential quadratic programming methods~\cite{zhuSequentialQuadraticProgramming2024}, iteratively update all players' strategies simultaneously. %
Further developments include iterative Linear-Quadratic Games (iLQGames)~\cite{fridovich-keilEfficientIterativeLinearQuadratic2020, rowold2024open} for fast feedback solutions and %
learning-based techniques for accelerated equilibrium computation \cite{petersLearningMixedStrategies2022}.

\smallskip\noindent\textbf{Race rules and experiment design.}
A naive design of race conditions and rules can make it difficult to study the performance gap between methods, because most races end in a draw or feature limited interaction.
To this end, prior work forces interaction by choosing asymmetric starting positions and velocities \cite{kangAutonomousMultidroneRacing2024},
handicapping leading players~\cite{spicaRealTimeGameTheoretic2018,linigerNonCooperativeGameApproach2019}, or asymmetrically assigning collision-avoidance responsibility~\cite{thakkarHierarchicalControlHeadtoHead2024,jiaRAPIDAutonomousMultiAgent2023}.

\smallskip\noindent\textbf{Connection to this work.}
In this work, we design racing rules and experiment conditions that yield highly competitive and interactive scenarios.
Our racing rules and conditions are inspired by \cite{spicaRealTimeGameTheoretic2018} but extend to velocities five times greater than in that prior work.
To handle these high-speed competitive settings, we develop LMPG, which builds upon~\cite{petersLearningMixedStrategies2022}.
We compare our LMPG approach against a contouring MPC formulation akin to~\cite{romeroModelPredictiveContouring2022} and a generalized Nash MPG solver akin to the one proposed in~\cite{cinar2025does}.

\section{Racing formulation}

The autonomous race is structured as a competitive multi-agent scenario, where each player aims to navigate the race track while optimizing its own control strategy under a set of predefined rules. The race environment consists of a closed-loop race track with a known layout, parameterized using a smooth periodic spline representation. The race track includes designated checkpoints in the form of gates as shown in \figref{fig:motivation}. These gates serve as verification points that enforce adherence to the intended trajectory, preventing shortcuts or excessive deviations from the track.

\begin{table*}[t]
\caption{Formalized racing rules designed to ensure fair head-to-head competition and enforce safety boundaries.}
\label{tab:racing_rules}
\begin{tabularx}{\textwidth}{clX}
\toprule
\textbf{ID} & \textbf{Rule} & \textbf{Description} \\
\midrule
R1 & Role assignment & Players are assigned attacker (behind) or defender (in front) roles. Roles switch after a valid overtake. \\
R2 & Overtaking & A valid overtake occurs when the attacker is \SI{0.75}{\meter} ahead of the defender. \\
R3 & Collision responsibility & The attacker is responsible for collision avoidance and must maintain a \SI{0.35}{\meter} distance ($r_{\text{col}}$) from the defender. \\
R4 & Gate passage & All gates must be passed through. \\ 
R5 & Track limits & Players must not deviate more than \SI{2}{\meter} off-track. \\ 
R6 & Velocity limits & Players must respect role-dependent speed limits. \\ 
R7 & Race duration & The race is limited to a maximum of \SI{5} laps. \\
R8 & Winner determination & The winner is the player with the most time spent as defender (leading). This incentivizes continuous competition for the lead and prevents passive last-lap overtaking strategies. Violating any safety constraint or operating limits (R3-R6) results in a loss for the offending agent. \\
\bottomrule
\end{tabularx}
\vspace{-1em}
\end{table*}

\subsection{Assumptions}
The focus of this paper is on control and planning, rather than perception. To simplify the problem, we assume:
\begin{enumerate}
\item \textbf{Full knowledge of opponents' states:} each player has complete information about the current states of both players at any given time, including their positions, velocities, and accelerations.
\item \textbf{Full global knowledge of the race track:} each player knows the layout of the race track, including its current position and progress along the track. 
\item \textbf{Non-cooperative setting:} the game is non-cooperative, meaning players do not share their strategies or intentions with each other. 
\end{enumerate}

\subsection{Racing Rules}
We employ a set of racing rules as summarized in \tabref{tab:racing_rules} that seek to mirror the high-level regulations of standard competitions, ensuring continuous progress and preventing stalemates. Each rule is motivated by potential ambiguities and practical considerations observed in autonomous racing. %

To evaluate whether agents comply with these rules, we implement an asynchronous referee system that monitors player behavior and determines race events such as the start, finish, and disqualifications. Notably, the referee also determines the current attacker and announces it to all the agents. It operates only on the positions and velocities of all players without any access to their strategies.

\section{Model Predictive Techniques for Autonomous Racing}
\label{sec:main}

We formulate the head-to-head racing task as a two-player non-cooperative dynamic game played over a receding horizon of length $K$ with stage $k \in \mathcal{K} = \{0, \dots, K-1\}$. At any instance, player~$i$ must compute a control sequence $\boldsymbol{u}^i = \{ \boldsymbol{u}^i_0, \boldsymbol{u}^i_1, \dots, \boldsymbol{u}^i_{K-1}\}$ that maximizes its racing performance.

\smallskip\noindent\textbf{Remark on notation:}
Throughout this document we use superscripts to index players and subscripts to index time.
Omission of one of these indices denotes aggregation over that dimension.
For example,  $\boldsymbol{u}^i$ denotes the input sequence for player~$i$ over the horizon,\BLUE{~$\neg i$ denotes player~$i$'s opponent}, and $\boldsymbol{u}$ denotes the input sequence for \emph{all} players jointly.

\subsection{General Problem Formulation: Model Predictive Game}
Due to the non-cooperative nature of the task, each agent's optimal strategy depends on their opponent's actions.
We therefore cast the racing problem as a Nash Equilibrium Problem, i.e., a coupled optimization problem in which each agent's strategy is the best response to the other's ~\cite{facchinei2010generalized}. 

In this setting, the planner solves for an equilibrium profile of trajectories $((\textcolor{mpc_color}{\boldsymbol{x}^{i*}}, \textcolor{mpc_color}{\boldsymbol{u}^{i*}}), (\textcolor{mpg_color}{\boldsymbol{x}^{\neg i*}}, \textcolor{mpg_color}{\boldsymbol{u}^{\neg i*}}))$.
At such an equilibrium, no players can reduce their cost by unilaterally deviating, i.e. $\forall i \in \{1,2\}$:  
\begin{equation} \label{eq:ocp_mpg}
    \begin{aligned}
        \underbrace{\textcolor{mpc_color}{\boldsymbol{x}^{i*}}, \textcolor{mpc_color}{\boldsymbol{u}^{i*}}}_{\pi^i_{\text{MPG}}(\boldsymbol{x}_{\text{init}})} & \in \underset{\boldsymbol{x}^i, \boldsymbol{u}^i}{\mathrm{argmin}} \enspace J^i(\boldsymbol{x}^i, \boldsymbol{u}^i, \textcolor{mpg_color}{\boldsymbol{x}^{\neg i*}}) \\[-1em]
        & \begin{aligned}
            \mathrm{s.t.}  \quad & \left. \begin{aligned}
            \boldsymbol{x}_{k+1}^i &= \boldsymbol{f}_k(\boldsymbol{x}^i_k, \boldsymbol{u}^i_k) \\
             h^i(\boldsymbol{u}^i_k) &\geq 0
            \end{aligned}\right\}\forall k\in\mathcal{K}\\
             & \quad\quad\boldsymbol{x}^i_0= \boldsymbol{x}^i_{\text{init}}
        \end{aligned}
    \end{aligned}
\end{equation}
Here, $J^i$ denotes the agent-specific objective function (detailed in Sec. \ref{sec:cost}), and $\boldsymbol{h}^i$ enforces the actuation limits. 
Here, $\boldsymbol{f}_k$ and $\boldsymbol{x}_{\text{init}}$ refer to the discretized system dynamics and current initial state, respectively.
Note that we solve Problem (\ref{eq:ocp_mpg}) in a decentralized manner. 
An agent using MPG solves this problem to generate two artifacts simultaneously: $\boldsymbol{x}^{\neg i*}$ serves as a game-theoretic prediction of the opponent, and $\boldsymbol{x}^{i*}$ is the corresponding best response.
By solving for this equilibrium problem, the MPG strategy inherently captures the coupled nature of agents' decision-making.

\smallskip\noindent\textbf{Remark on practical realization:}
In practice, solving Problem (\ref{eq:ocp_mpg}) to a global solution is intractable~\cite{cinar2025does, lecleachALGAMESFastAugmented2022, zhuSequentialQuadraticProgramming2024}.
Therefore, like these prior works, we only seek local equilibria by solving for the first-order necessary conditions of~(\ref{eq:ocp_mpg}) via the PATH solver~\cite{dirksePathSolverNommonotone1995}.

\begin{figure*}[t!]
    \centering
    \includegraphics[width=\linewidth]{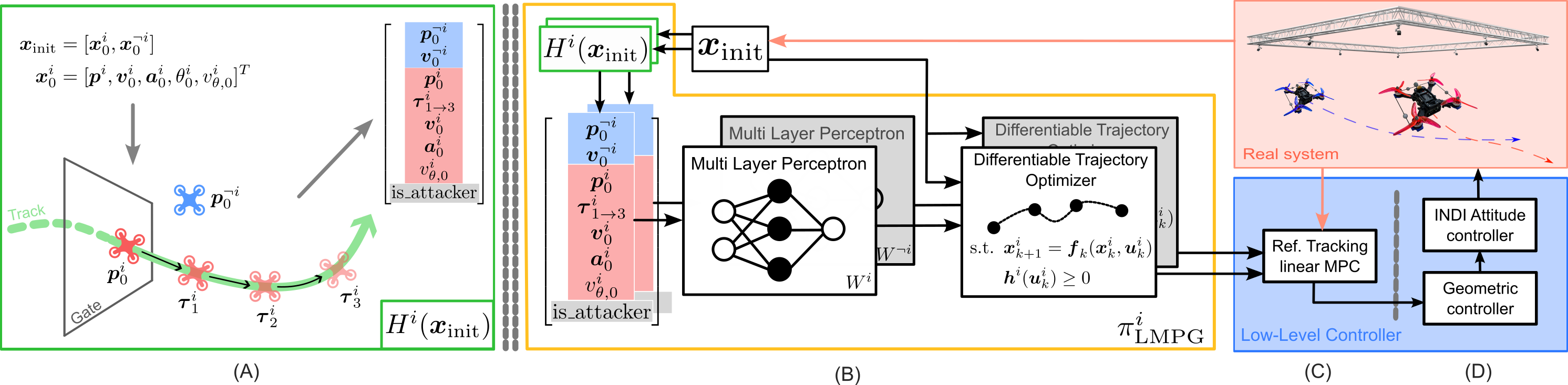}
    \caption{
    (A) Composition of the observations used by the neural network employed by LMPG~(c.f.~\ref{eq:lmpg_input}).
    (B) Pipeline overview of LMPG: observations are fed into a neural network that embeds differentiable trajectory optimization as a final layer to predict a receding-horizon strategy. 
    (C) A linear MPC tracks the reference strategy via feedback linearization. In simulation, the output of this controller is directly fed into the simulated dynamics.
    (D) Hierarchical control scheme employed in hardware experiments: using the Agilicious framework~\cite{Foehn_2022}, the drone tracks the receding-horizon strategy by combining a geometric controller with incremental nonlinear dynamic inversion (INDI)~\cite{sun2022comparative}.
    }
    \label{fig:lmpg_diagram}
    \vspace{-1em}
\end{figure*}

\subsection{The Predict-Then-Plan MPC Baseline}
Solving for the equilibrium condition in \eqref{eq:ocp_mpg} is computationally demanding due to the coupling between players.
To address this, a standard %
approximation is to treat the opponent not as a strategic agent, but as a dynamic obstacle with a fixed, non-reactive strategy.
This approximation reduces the coupled optimization problem in \eqref{eq:ocp_mpg} to a single-player optimization problem following an MPC baseline:
\begin{equation} \label{eq:ocp_mpc}
    \begin{aligned}
        \underbrace{\textcolor{mpc_color}{\boldsymbol{x}^{i*}}, \textcolor{mpc_color}{\boldsymbol{u}^{i*}}}_{\pi^i_{\text{MPC}}(\boldsymbol{x}_{\text{init}})} \in \enspace & \underset{\boldsymbol{x}^i, \boldsymbol{u}^i, \boldsymbol{x}^{\neg i}}{\mathrm{argmin}} \enspace J^i(\boldsymbol{x}^i, \boldsymbol{u}^i, \boldsymbol{x}^{\neg i}) \\[-1em]
        \mathrm{s.t.} \enspace  & \left.
        \begin{aligned}
        \boldsymbol{x}^i_{k+1} &= \boldsymbol{f}_k(\boldsymbol{x}^i_k, \boldsymbol{u}^i_k) \\
        \boldsymbol{h}^i(\boldsymbol{u}_k^i) &\geq 0 \\
        \end{aligned}
        \right\} \forall k \in \mathcal{K}  \\
        & \enspace\quad\boldsymbol{x}^{\neg i} = \boldsymbol{F}_{\text{pred}}(\boldsymbol{x}^{\neg i}_0) \\
        & \quad\quad \boldsymbol{x}^i_0 = \boldsymbol{x}^i_{\text{init}} 
    \end{aligned}
\end{equation}
In this formulation, $\boldsymbol{F}_{\text{pred}}$ is a prediction model that governs the evolution of the opponent's state.
In Sec.~\ref{sec:sim_res}, we will consider two different prediction models to instantiate variants of this baseline that capture different trade-offs between computational complexity and prediction accuracy.
While this approximation significantly reduces computational complexity, it neglects the opponent's reactive capabilities, potentially leading to suboptimal behavior in highly interactive scenarios.

\subsection{Planner Dynamics Model}
To define the system dynamics $\boldsymbol{f}_k$ used by MPC and MPG, we adopt a simplified point mass model with jerk control.
This serves as an abstraction of the full vehicle dynamics for the planner.
The state $\boldsymbol{x}^i$ and input $\boldsymbol{u}^i$ are:
\begin{subequations} \label{eq:dynamics_player_i}
    \begin{align}
        \boldsymbol{x}^i &= [p_x \enspace p_y \enspace p_z \enspace v_x  \enspace v_y  \enspace v_z \enspace a_x \enspace a_y \enspace a_z \enspace \theta \enspace v_{\theta}]^T, \quad  \\
        \boldsymbol{u}^i &= [j_x \enspace j_y \enspace j_z \enspace \Delta v_\theta]^T \label{eq:state_and_input_space}
    \end{align}
\end{subequations} 
Here, $p_{(\cdot)}$, $v_{(\cdot)}$, and $a_{(\cdot)}$ denote the Cartesian position, linear velocity, and acceleration components in the inertial frame, respectively. The variable $\theta$ represents the progress along the track centerline, while $v_{\theta}$ is the progress velocity. The control inputs consist of the linear jerk components $\boldsymbol{j} = [j_x, j_y, j_z]^T$ and the commanded path acceleration $\Delta v_\theta$. 
We obtain $f_k$ by discretizing the continuous dynamics $\dot{\boldsymbol{x}}^i = [\boldsymbol{v} \enspace \boldsymbol{a} \enspace \boldsymbol{j} \enspace v_\theta \enspace \Delta v_\theta]^T$  using a zero-order hold approximation.

\subsection{Racing Cost Function} \label{sec:cost}
The cost function $J^i$ drives the racing behavior in both the MPG and MPC formulations. Over the prediction horizon $K$, this objective balances track progress against race-line tracking accuracy and control effort. To ensure the solver finds a solution even in constrained scenarios, strict safety requirements such as collision avoidance and velocity limits are implemented as soft penalty terms.

The cost $J^i$, inspired by~\cite{romeroModelPredictiveContouring2022}, is defined as:
\begin{subequations} \label{eq:mpg_cost_player_i}
    \begin{align}
    J^i(\cdot)
    = & \sum_{k=0}^{K-1} \eqnmarkbox[mpc_color]{lit1}{ \enspace || e^l(\theta^i_{k})||^2_{q_l} + || e^c(\theta^i_{k})||^2_{q_c(\boldsymbol{p}^d(\theta_k^i))}} \label{eq:mpg_cost_player_i_a} \\
    & \eqnmarkbox[mpc_color]{lit2}{+ || \boldsymbol{u}^{i}_k|| ^2_{q_u}} \label{eq:mpg_cost_player_i_d} \\
    & \eqnmarkbox[mpc_color]{lit3}{+ \mu \cdot (v^{\neg i}_{\theta, k} - v^i_{\theta, k})} \label{eq:mpg_cost_player_i_e} \\
    & + \boldsymbol{1} \{ \mathrm{is\_attacker} = i\} \cdot q_{\text{col}} \cdot \Omega(\boldsymbol{p}^i_k, \boldsymbol{p}^{\neg i}_k, r_{\mathrm{col}})
    \label{eq:mpg_cost_player_i_b} \\
    & + q_{\mathrm{vel}} \cdot \Psi(||\boldsymbol{v}^i||, v_{\mathrm{max}}) \label{eq:mpg_cost_player_i_c} 
    \end{align} 
\end{subequations}
\annotate[yshift=-0.5em]{below,right}{lit1}{As in \cite{romeroModelPredictiveContouring2022}} 
The terms in \eqref{eq:mpg_cost_player_i_a} minimize contouring error $e^c$ and lag error $e^l$ relative to the reference path $\boldsymbol{p}^d$. 
The term in \eqref{eq:mpg_cost_player_i_d} acts as a regularization term on the input while \eqref{eq:mpg_cost_player_i_e} encourages the agent to maximize its own progress velocity $v^i_{\theta}$ relative to the opponent weighted by a scalar parameter $\mu$, facilitating overtaking \cite{wangMultiagentSensitivityEnhanced2020}. 
\eqref{eq:mpg_cost_player_i_b} represents the collision cost $\Omega$, active only when the agent is the attacker and within a proximity radius $r_{\mathrm{col}}$ of the opponent.
Finally, \eqref{eq:mpg_cost_player_i_c} imposes a soft penalty $\Psi$ for violating the maximum velocity $v_{\mathrm{max}}$, which varies depending on whether the agent is attacking or defending. 
This design is consistent with the racing rules established earlier, which impose the responsibility for collision avoidance on the overtaking drone while admitting it a higher maximum velocity.

\section{Learned Model Predictive Game}

\begin{figure*}[t!]
    \centering 
    \begin{subfigure}[b]{0.24\textwidth}
        \centering
        \includegraphics[width=1.0\linewidth]{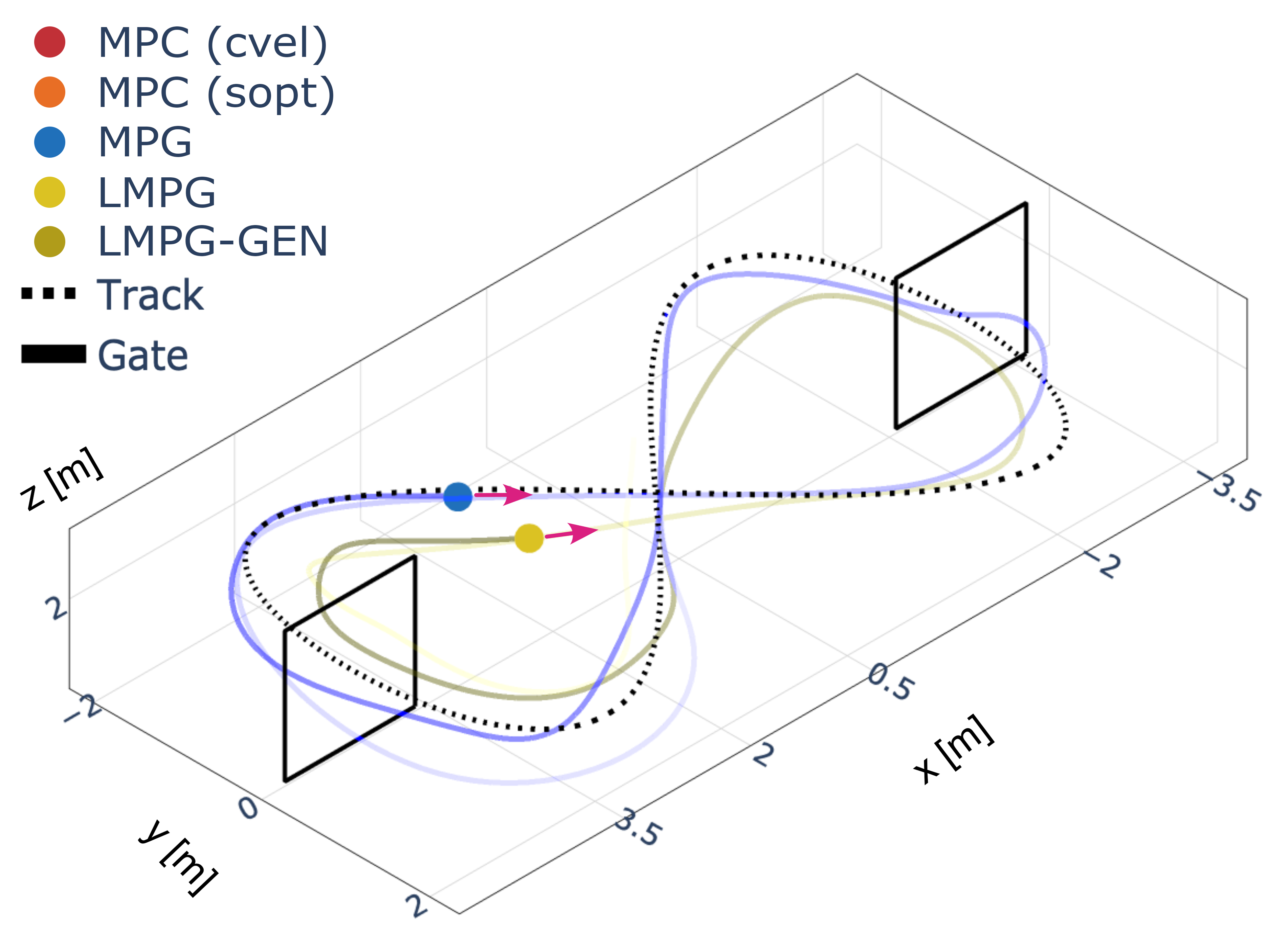}
        \caption{Lemniscate track}
        \label{fig:track_lemniscate}
    \end{subfigure}    
    \begin{subfigure}[b]{0.24\textwidth}
        \centering
        \includegraphics[width=1.0\linewidth]{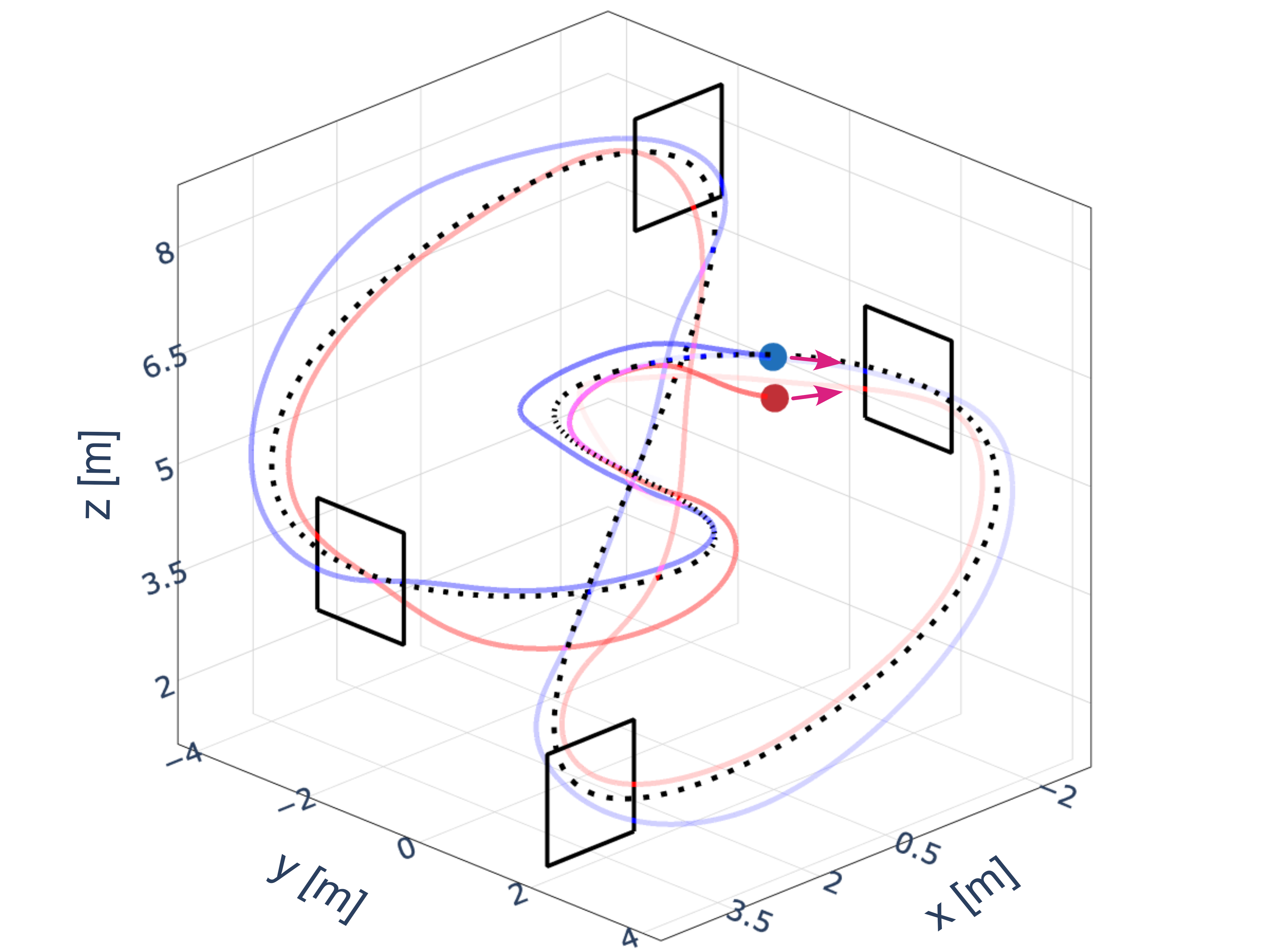}
        \caption{Lissajous track}
        \label{fig:track_lissajous} 
    \end{subfigure}
    \begin{subfigure}[b]{0.24\textwidth}
        \centering
        \includegraphics[width=1.0\linewidth]{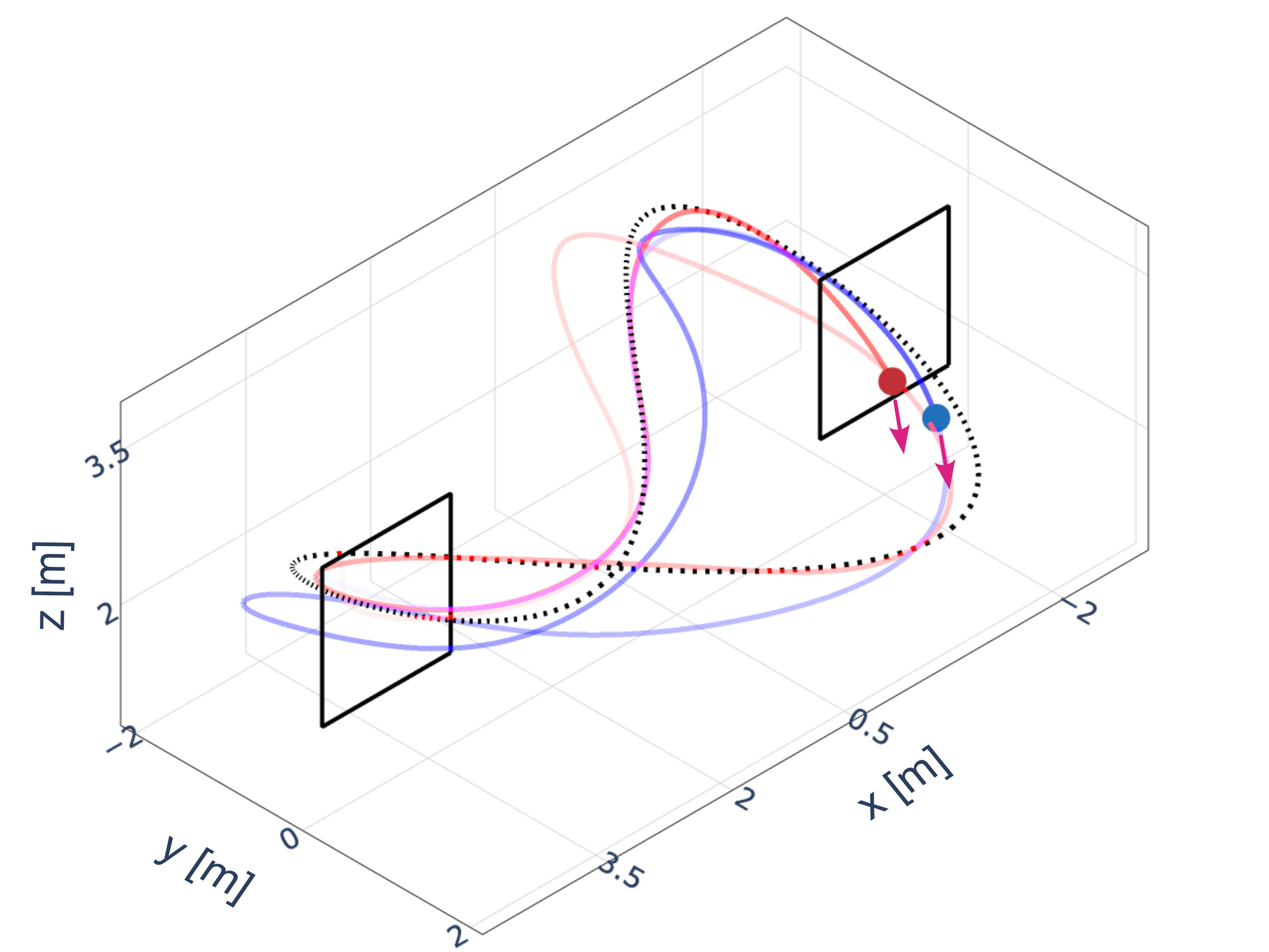}
        \caption{3D Lemniscate track}
        \label{fig:track_eight}
    \end{subfigure}
     \begin{subfigure}[b]{0.24\textwidth}
        \centering
        \includegraphics[width=1.0\linewidth]{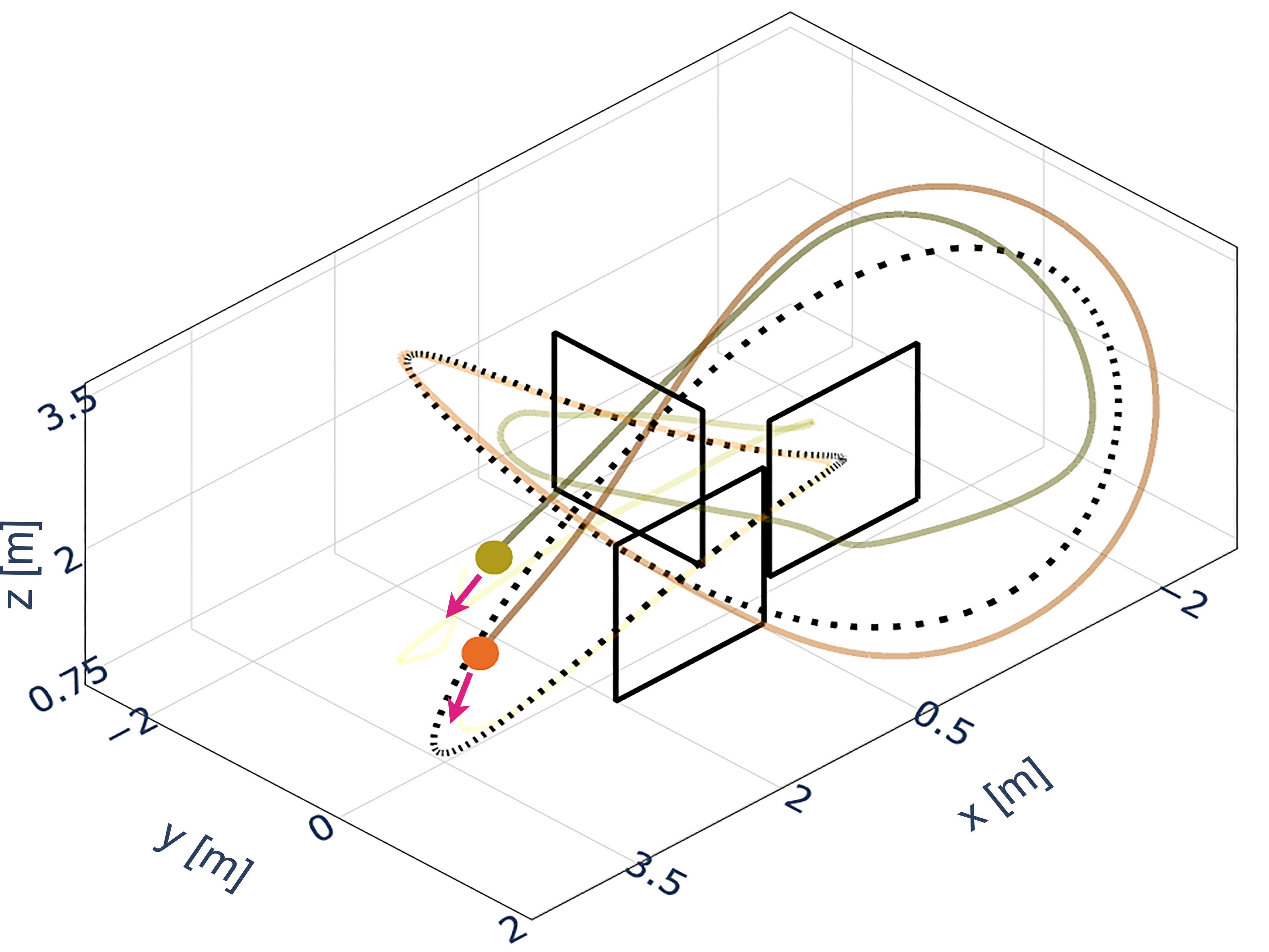}
        \caption{Trefoil track}
        \label{fig:track_trefoil}
    \end{subfigure}
    \caption{The three race tracks used for experimental evaluation: (a) Lemniscate, (b) Lissajous, (c) 3D Lemniscate, and (d) Trefoil}. The colored lines show sample trajectories from the head-to-head tournament, visualizing overtaking maneuvers between the competing methods.
    \vspace{-1em}
    \label{fig:tracks}
\end{figure*}

To accelerate the decision-making process of MPG, we propose the Learned Model Predictive Game (LMPG).
LMPG amortizes the computation of receding-horizon game solutions in~\eqref{eq:ocp_mpg} by training a structured neural network $\pi^{W^i}_\text{LMPG}$ for each player~$i$ that \emph{predicts} their Nash strategy given the current state.

\subsection{Amortized Game Formulation}
To achieve this amortization, we translate the game of~\eqref{eq:ocp_mpg} into a game over each player's neural network parameters $W^i$, i.e.,
\begin{equation} \label{eq:ocp_lmpg}
    \begin{aligned}
        W^{i*} & \in\underset{W^i}{\mathrm{argmin}}\scalebox{0.9}{$\displaystyle\enspace \underset{\boldsymbol{x}_{\text{init}} \sim D}{\mathbb{E}} \Big[ J^i \big (\pi^{W^{i}}_{\text{LMPG}}(\boldsymbol{x}_{\text{init}}), \pi^{W^{\neg i*}}_{\text{LMPG}}(\boldsymbol{x}_{\text{init}}) \big) \Big]$}
    \end{aligned}
\end{equation}
In this game, each player seeks to minimize their \emph{expected} receding-horizon cost over a dataset $\mathcal{D}$ of initial states. We emphasize that $J^i$ represents the same strategic cost function defined in~\eqref{eq:ocp_mpg}.

\smallskip\noindent\textbf{LMPG policy structure.}
\figref{fig:lmpg_diagram}b summarizes the special structure of the LMPG policy, $\pi^i_\text{LMPG}$.
As shown, two multilayer perceptrons (MLPs) predict reference inputs $\boldsymbol{\hat{u}}^i$ based on an observation of the current initial state $\boldsymbol{x}_\text{init}$.
To ensure feasibility, this policy includes a differentiable trajectory-optimization layer finds the closest strategy that adheres to dynamics and input constraints, $\boldsymbol{f}$ and $\boldsymbol{h}^i$:
\begin{equation}
    \begin{aligned}
        \underbrace{\textcolor{lmpg_color}{\boldsymbol{x}^{i*}}, \textcolor{lmpg_color}{\boldsymbol{u}^{i*}}}_{\pi^i_{\text{LMPG}}(\boldsymbol{x}_{\text{init}})} \in \enspace & \underset{\boldsymbol{x}^i, \boldsymbol{u}^i}{\mathrm{argmin}} 
        \sum_{k=0}^{K-1}|| \boldsymbol{u}_k^i - \boldsymbol{\hat{u}}_k^i||^2_2 \\[-1em]
        \mathrm{s.t.} \enspace &   \left.
        \begin{aligned}
        \boldsymbol{x}^i_{k+1} &= \boldsymbol{f}_k(\boldsymbol{x}^i_k, \boldsymbol{u}^i_k) \\
        \boldsymbol{h}^i(\boldsymbol{u}_k^i) &\geq 0 \\
        \end{aligned}
        \right\} \forall k \in \mathcal{K}  \\
        & \quad\quad \boldsymbol{\hat{u}^i} = \text{MLP}^{W^i} \big( H^i(\boldsymbol{x}^{i}_{\text{init}}) \big) \\
        & \quad\quad \boldsymbol{x}^i_0 = \boldsymbol{x}^i_{\text{init}}
    \end{aligned}
\end{equation}

\noindent\textbf{Observation encoding.}
To ensure generalization, the MLP does not process raw initial states~$\boldsymbol{x}_\text{init}$ directly but instead observes postprocessed observations,
\begin{equation} \label{eq:lmpg_input}
    H^i(\boldsymbol{x}_\text{init}) = \scalebox{0.9}{$\displaystyle [
    \boldsymbol{p}^{\neg i}_0, \boldsymbol{v}^{\neg i}_0,
    \boldsymbol{p}^i_\mathrm{0}, \boldsymbol{\tau}^i_\mathrm{1 \rightarrow 3},
    ,\boldsymbol{v}^i_0,  \boldsymbol{a}^i_0, v^i_{\theta,0}, \mathrm{is\_attacker}]^T $}
\end{equation}
This observation encoding includes three equidistant reference points spaced \SI{0.75}{\meter} apart along the race-track $\boldsymbol{\tau}^i_{1 \rightarrow 3} = [\boldsymbol{\tau}^i_1, \boldsymbol{\tau}^i_2, \boldsymbol{\tau}^i_3]$ as well as the opponent's position $\boldsymbol{p}^{\neg i}_0$  and velocity $\boldsymbol{v}^{\neg i}_0$, all expressed in the body frame of agent $i$ as shown in~\figref{fig:lmpg_diagram}a. The remaining terms $\boldsymbol{p}^i_\mathrm{0},\boldsymbol{v}^i_0,  \boldsymbol{a}^i_0, v^i_{\theta,0}$ refer to agent $i$'s current kinematic state and strategic role determined by the $\mathrm{is\_attacker}$ boolean variable.

\subsection{Training Procedure}
To solve the game over neural-network weights in~\eqref{eq:ocp_lmpg}, we adopt a simultaneous gradient play approach, where each agent's network parameters $W^i$ are updated iteratively via
\begin{align}\label{eq:lmpg_update}
        &W^i \gets W^i - \alpha \delta W^i \text{\quad where} \\
        &\delta W^i = \nabla_{W^i} J^i \Big(\pi^{W^{i}}_{\text{LMPG}}(\boldsymbol{x}_{\text{init}}), \pi^{W^{\neg i}}_{\text{LMPG}}(\boldsymbol{x}_{\text{init}}) \Big)\nonumber
\end{align}
where the gradient $\delta W^i$ is obtained via automatic differentiation. 
To compute gradients through the optimization layer $\mathcal{P}$, we employ the differentiable optimization machinery proposed by \cite{petersLearningMixedStrategies2022}, which allows for efficient backpropagation through the Karush-Kuhn-Tucker (KKT) conditions of the trajectory optimizer. 
The update minimizes the respective cost functions $J^i$ over initial states $\boldsymbol{x}_{\text{init}}$ sampled from a dataset $\mathcal{D}$, at a learning rate~$\alpha$.
Algorithm~\ref{alg:lmpg_training} describes our training procedure employing this update rule.
Rather than applying~\eqref{eq:lmpg_update} to a dataset of randomly sampled initial states, this training procedure employs a specialized data aggregation strategy to facilitate stable, generalizable learning.
We summarize this data aggregation strategy below.

\smallskip\noindent\textbf{Data aggregation strategy.}
Each training epoch begins with a data collection phase (lines~\ref{line:rollout_sim_delays}-\ref{line:end_rollout_sim_delays} of ~Alg.~\ref{alg:lmpg_training}) in which we roll out the current learned strategies $\pi^i_\text{LMPG}$ in a closed-loop simulation from randomly initialized game states.
To bridge the sim-to-real gap, we inject randomized computational delays from a Bernoulli distribution, $\mathrm{Bernoulli}$, and control input noise from a normal distribution~$\mathcal{N}$ during these rollouts, sensitizing the policy to the asynchronous nature of real-life deployment.
\BLUE{Furthermore, the differentiable trajectory optimization's constraints are relaxed during data collection to encourage exploration. Specifically, we introduce a relaxation margin $\boldsymbol{\epsilon} > 0$, replacing the strict constraint $\boldsymbol{h}^i(\boldsymbol{u}_k^i) \geq 0$ with the relaxed condition $\boldsymbol{h}^i(\boldsymbol{u}_k^i) \geq -\boldsymbol{\epsilon}$.}

\begin{algorithm}[t]
\caption{LMPG Training Procedure}
\label{alg:lmpg_training}
\begin{algorithmic}[1]
\Require learning rate $\alpha$, max epochs $E$, max episodes $M$
\State \textbf{Initialize:} Policy parameters $W^i$ for all players $i \in {1,2}$
\For{epoch $= 1$ to $E$}
    \State{Empty dataset $\mathcal{D}$}
    \For{episode $= 1$ to $M$}
        \State \textbf{//Rollout with simulated delays} \label{line:rollout_sim_delays}
        \State Sample random initial state $\boldsymbol{x}_{\text{init}}$
        \State \textbf{Initialize:} $\boldsymbol{u}^{i*}_{\text{prev}}, k^i \gets (\mathbf{0}, 0)$ for each player $i$
        \State Store $\boldsymbol{x}_{\text{init}} = (\boldsymbol{x}^1_{\text{init}}, \boldsymbol{x}^2_{\text{init}})$ in $\mathcal{D}$
        \While{race not finished}
            \For{each player $i$}
                \State $(\boldsymbol{x}^{i*}, \boldsymbol{u}^{i*}) \gets \pi^i_{\text{LMPG}}(\boldsymbol{x}_{\text{init}})$
                \State Sample delay mode $d^i \sim \mathrm{Bernoulli}(0.5)$
                \State Sample random noise $\nu^i \sim \mathcal{N}(0, \nu_{max})$

                \If{$d^i=1$} \Comment{Use previous strategy}
                    \State $\boldsymbol{u}^{i*}, k^i \gets (\boldsymbol{u}^{i*}_{\text{prev}}, k^i +  1)$
                \Else \Comment{No delay: use current strategy}
                    \State Apply noise: $\boldsymbol{u}^{i*} \leftarrow \boldsymbol{u}^{i*} + \nu^i$
                    \State Update memory: $\boldsymbol{u}^{i*}_{\text{prev}}, k^i \gets (\boldsymbol{u}^{i*}, 0)$
                \EndIf
                Step sim: $\boldsymbol{x}^i_{\mathrm{init}} \leftarrow \mathrm{MPC}_\text{lin}(f_k(\boldsymbol{x}^i_{\text{init}}, \boldsymbol{u}^{i*}_{k^i}))$
            \EndFor
            \State Store $\boldsymbol{x}_{\text{init}} = (\boldsymbol{x}^1_{\text{init}}, \boldsymbol{x}^2_{\text{init}})$ in $\mathcal{D}$ \label{line:end_rollout_sim_delays}
        \EndWhile
    \EndFor
    \State \textbf{//Simultaneous gradient play}
    \State{Shuffle dataset $\mathcal{D}$}
    \For{$\boldsymbol{x}_{\text{init}}$ in $\mathcal{D}$}
         \State Update parameters with \eqref{eq:lmpg_update} for each player $i$
    \EndFor
\EndFor
\end{algorithmic}
\vspace{-0.25em}
\end{algorithm}

\section{Simulation Results} \label{sec:sim_res}
\begin{figure*}[t!]
    \centering
    \includegraphics[width=\linewidth]{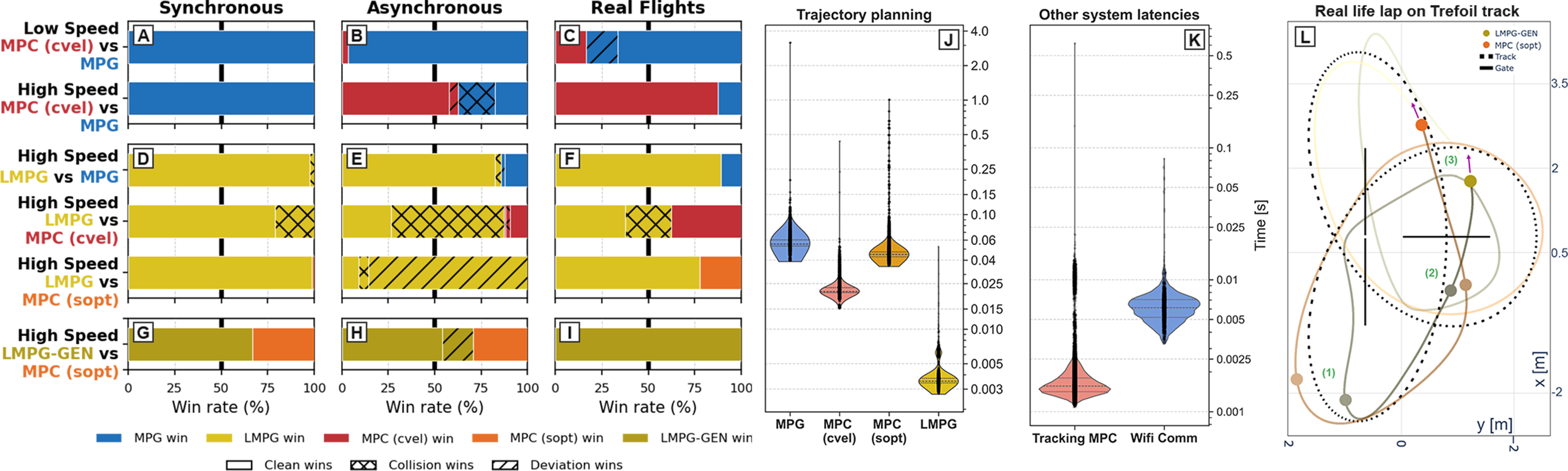}
    \caption{\textbf{Head-to-head racing results in simulation and real-life.} 
    \textbf{(A, B)} Win rates for MPC (cvel) vs. MPG in a simulated tournament with synchronous (A) and asynchronous (B) execution modes.
    \textbf{(D, E)} Win rates for LMPG vs. MPC/MPG in a simulated tournament with synchronous (D) and asynchronous (E) execution modes.
    \textbf{(G, H)} Zero-shot win rates for LMPG-GEN vs. MPC (sopt) in a simulated tournament with synchronous (G) and asynchronous (H) execution modes.
    \textbf{(C, F)} Win rates for the real-world flight tournament on the lemniscate track.
    \textbf{(I)} Zero-shot win rates for the real-world flight tournament on the trefoil track.
    \textbf{(J)} Trajectory planning time distributions for all methods.
    \textbf{(K)} Low level system latencies distributions encountered during real-life experiments.
    \textbf{(L)} Vizualization of a single real-life lap of LMPG-GEN deployed zero-shot vs MPC (sopt) on the trefoil track.
    }
    \vspace{-1em}
    \label{fig:paper_banner_results}
\end{figure*}

\label{sec:exp}
Our simulation experiments are designed to rigorously evaluate the performance of LMPG, MPG, and two MPC variants that differ in their prediction model $F_\mathrm{pred}$ of \eqref{eq:ocp_mpc}.
The first variant, MPC (cvel), makes a constant velocity prediction \BLUE{for the opponent, akin to~\cite{kaufmann2023champion}}.
The second variant, MPC (sopt), predicts the opponent's strategy that minimizes the contouring cost~(4) subject to dynamics and input constraints, yielding a more accurate prediction sensitive to track geometry and single-agent constraints, \BLUE{similar to~\cite{rowoldEfficientSpatiotemporalGraph2022}}.
Since the ego agent's strategy is not known during this prediction problem, we drop the collision avoidance term in Equation~(4e) and fix $v_{\theta}^{\neg i}$ at the current velocity.

All of these methods have vastly different characteristics with respect to their solver latency~(c.f., \figref{fig:paper_banner_results}j) and strategic reasoning capabilities: %
MPG computes a high-fidelity strategy but suffers from high solve times;
MPC~(cvel) and MPC (sopt) compute solutions more quickly at the expense simplified opponent prediction models;
LMPG approximates the high-fidelity strategy of MPG while matching the solve times of MPC (cvel).
Therefore, a comparison of these methods allows us to study the trade-off between these characteristics.

We present our results as follows: first in Sec.~\ref{sec:exp-sync} to ~\ref{sec:exp-async}, we compare the online planners at the extreme ends of the design spectrum, MPC~(cvel) vs. MPG.
We study these methods under varying execution modes to assess the impact of solver latency and prediction accuracy on closed-loop racing performance.
In Sec.~\ref{sec:exp-accelerated}  we compare LMPG against all model-based methods.
Finally, Sec.~\ref{sec:exp-generalized} analyses on the generalization capabilities of LMPG. 

\subsection{Experiment Setup}\label{sec:exp-setup}
We implement a structured tournament format where each method competes against others in a round-robin style.
All of these experiments share the following experimental setup:

\smallskip\noindent\textbf{Race tracks.}
A tournament is played on three distinct race tracks, visualized in \figref{fig:tracks}a-c.
Parameterized by periodic cubic splines, these tracks are designed to introduce unique challenges in terms of layout complexity, gate configurations, and required maneuvering skills.

\smallskip\noindent\textbf{Speed configurations.}
We consider two velocity configurations.
The \emph{low speed} setting is designed to replicate velocity ratios from~\cite{spicaRealTimeGameTheoretic2018}, with a maximum attacker speed of \SI{2}{\meter\per\second}.
The \textit{high speed} setting presents a more challenging configuration with a maximum attacker speed of \SI{3}{\meter\per\second}.
In both settings, the defender's maximum velocity is set \SI{1}{\meter\per\second} lower than the attacker's to encourage overtaking maneuvers.

\smallskip\noindent\textbf{Tournament structure.}
For each comparison group, we run a total of $120$ races ($40$ per track).
To study diverse racing conditions, we sample $20$ initial positions per track uniformly from spherical regions (\SI{0.15}{\meter} radius) behind the start line. %
For every sampled position, the pair of agents races twice, swapping roles and starting positions between attacker and defender. This ensures both agents experience the exact same initial conditions from both offensive and defensive perspectives.

\smallskip\noindent\textbf{Additional implementation details.}
All methods utilize the PATH solver~\cite{dirksePathSolverNommonotone1995} to solve the game equilibrium conditions formulated as a Mixed Complementarity Problem (MCP). 
Training typically concludes in approximately 3 hours over 600k gradient iterations.
Furthermore, the experiments follow the hierarchical control scheme from \figref{fig:lmpg_diagram}c, in which the planner’s output is refined by a reference-tracking linear MPC before being applied to the drone. This low-level control block is implemented to run independently of the planner and is not bound by the planner's solve time. 

\subsection{How do MPC (cvel) and MPG perform if computation time is negligible?}
\label{sec:exp-sync}

\smallskip\noindent\textbf{Execution mode: synchronous.}
Our first set of experiments instantiates the tournament in a simulation where the clock pauses during the agents' computation phase.
At each discrete timestep, all agents receive the current state and calculate their strategies taking as much wall-clock time as necessary before simultaneously executing their actions.
The environment advances only after all agents complete their calculations.
Consequently, this \emph{synchronous} execution mode strictly rules out the effect of computational delay.

\smallskip\noindent\textbf{Results.}
Fig.~\ref{fig:paper_banner_results}a shows the results of the tournament played in synchronous execution mode.
In this setting, we find that MPG consistently outperforms MPC (cvel) from \eqref{eq:ocp_mpc}, by achieving a perfect win rate across both speed configurations. This confirms the superiority of game-theoretic planning when computational constraints are removed. Furthermore, MPC (cvel) fails to execute a single overtake against MPG in synchronous trials. This disparity stems from the predictive models used by each solver: MPG leverages a game-theoretic model to anticipate optimal opponent responses, whereas MPC (cvel) utilizes a naive prediction model that assumes the opponent follows a fixed constant velocity trajectory.

\subsection{How do MPC (cvel) and MPG perform under fixed delays?}
\label{sec:exp-semi-sync}

\smallskip\noindent\textbf{Execution mode: synchronous with artificial delays.}
The compared planning methods exhibit vastly different timing characteristics, a critical factor ignored by the previous synchronous evaluation.
To isolate how this latency impacts strategic performance, independent of the variable solve times of each method, we introduce controlled artificial delays into the synchronous setting.
To this end, the simulation clock is stopped while agents compute their strategies (as in synchronous execution mode) but control inputs are executed at a fixed \emph{artificial delay}.
This execution mode therefore allows us to study the effect of latency on the methods' closed-loop performance in a controlled manner.

\begin{figure}[t!]
    \centering
    \includegraphics[width=\linewidth]{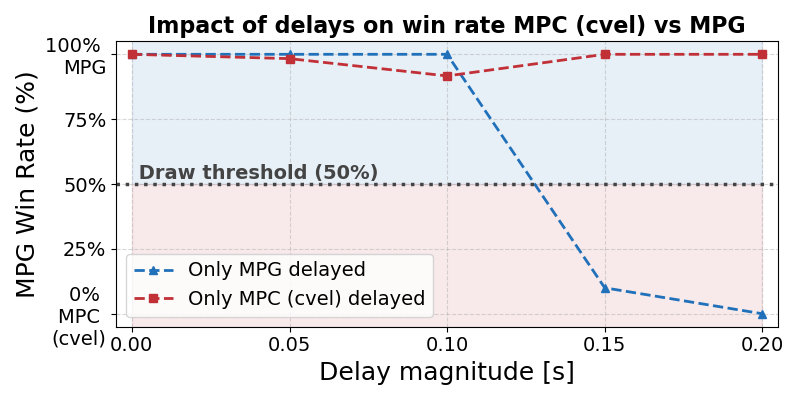}
    \caption{\textbf{Sensitivity of racing performance to latency.} 
    At negligible delay, MPG dominates MPC (cvel). As artificial delay is added to MPG's strategy (blue line), its performance degrades, allowing MPC (cvel) to win more races.}
    \vspace{-1em}
    \label{fig:delay_sensitivity}
\end{figure}

\smallskip\noindent\textbf{Results.}
\figref{fig:delay_sensitivity} presents the win rate of MPG against MPC (cvel) as a function of fixed delay magnitudes.
When both methods observe near-zero latency, MPG dominates as in the synchronous execution mode~(c.f. Sec.~\ref{sec:exp-sync})
When artificial delays are applied to MPG, MPC (cvel) secures more wins.
This result highlights that delays reduce the racing performance of MPG.

\subsection{How do MPC (cvel) vs MPG perform under solver-induced latency?}
\label{sec:exp-async}
\smallskip\noindent\textbf{Execution mode: asynchronous.}
In real-world racing, agents observe the environment, compute, and execute strategies at their own rates.
Hence, the execution delay depends on the varying solve times of each method  (c.f., \figref{fig:paper_banner_results}j).
We study this effect of variable delays induced by each method's computation time via an~\emph{asynchronous} execution mode, in which the simulation clock runs continuously and remains independent of the agents' reasoning time.
Therefore, if an agent incurs high computational latency, the environment state will have evolved by the time the action is applied, implicitly penalizing slow decision-making.

\smallskip\noindent\textbf{Results.}
Fig.~\ref{fig:paper_banner_results}b shows the results of the tournament played in asynchronous execution mode.
At low speeds, the physical dynamics are slow enough to mask the solver latency, allowing MPG to maintain its dominance. However, in the high-speed setting, the environment evolves too quickly for the solver to keep up. As visualized in the center and right panels of \figref{fig:paper_banner_results}, MPG loses its competitive edge, supporting our claim regarding the impact of computational overhead. While MPG averages a solve time of roughly \SI{60}{\milli\second}, complex interaction scenarios can cause spikes exceeding \SI{2}{\second}. This causes a violation where the drone lacks valid control input and drifts off-track.

In this setting, we observe that the computational overhead of MPG becomes a critical liability that effectively negates its strategic advantage.
However, the results also highlight a clear trade-off between safety and racing performance. Although MPG fails to finish frequently due to timeouts, it plans significantly safer overtakes, evidenced by the nonexistent collision wins for its opponent. In contrast, MPC (cvel)’s simplified prediction model allows for aggressive maneuvering but results in a significantly higher rate of self-inflicted collisions (seen in the cross-hatched portions of the bars).

Overall, these findings highlight a critical limitation of game-theoretic planning. While MPG remains superior in idealized synchronous conditions, its computational overhead becomes a severe bottleneck when real-time adaptability is required. These results underscore the necessity of optimizing solve times for game-theoretic methods to ensure their viability in real-time, high-speed racing scenarios.

\subsection{Can LMPG achieve a better trade-off between solve time and solution accuracy?}
\label{sec:exp-accelerated}

In a final set of simulation experiments, we  evaluate LMPG vs. MPC variants and MPG to validate the following claim: our learning-based game approximation retains the strategic benefits of MPG while eliminating computational bottlenecks.
To this end, we repeat the tournament in the high-speed configuration under the synchronous and asynchronous execution modes.

\smallskip\noindent\textbf{Results.}
\figref{fig:paper_banner_results}d-e show the results of these tournaments.
While MPG could only win when solver latency was negligible, LMPG overcomes this limitation: it reliably wins against MPC (cvel), MPG and MPC (sopt) in the most challenging high-speed setting even under asynchronous execution.
This performance can be attributed to two main factors.
First, offline training reduces inference time to \SI{3.5}{\milli\second} (a $14\times$ speedup, see \figref{fig:paper_banner_results}j), eliminating latency-induced failures and facilitating better reactivity.

Interestingly, we find that our amortized approach, LMPG, consistently outperforms the local solutions found by the non-amortized MPG method in the synchronous setting. While surprising, similar abilities to find better local minimizers than conventional approaches have been repeatedly reported in literature \cite{amos2023tutorial, sercu2021neural}.

\subsection{Does LMPG generalize across different tracks?}
\label{sec:exp-generalized}

While the relative track waypoint observation space is theoretically track-agnostic, the models in Sec.~\ref{sec:exp-accelerated} were trained for a specific track.
To test generalization, we train a single policy, LMPG-GEN, by aggregating experience from four morphologically diverse tracks and evaluate it (zero-shot) on the tracks of \figref{fig:tracks}) which are entirely unseen at training time.

\smallskip\noindent\textbf{Results.} We evaluated LMPG-GEN against MPC (sopt), which utilizes a more sophisticated opponent prediction model. The evaluation was conducted on the race tracks shown in~\figref{fig:tracks} which were entirely excluded from the training set. As shown in \figref{fig:paper_banner_results}g-h, LMPG-GEN consistently outperforms MPC (sopt) in both synchronous and asynchronous modes. Although we observed a marginal performance gap compared to the track-specific models, the generalized policy retains a significant competitive advantage over the predict-then-plan solver.

\section{Real-world Experiment Results}
To validate our simulation findings, we instantiate a variant of the tournament in a sequence of real-world experiments.
The primary goal of these experiments is to verify whether the trends observed in the high-speed asynchronous simulations (specifically, the performance degradation of MPG due to latency and the robustness of LMPG) hold under real-world aerodynamic and computational conditions.

\subsection{Do the simulation results transfer to real world?}
\label{sec:hw-results}

We adopt a hierarchical control scheme to bridge the gap between high-level planning and low-level actuation, as illustrated in \figref{fig:lmpg_diagram}d. This ensures robust and safe behavior during real-world flights.
The specific components of our hardware and software stack are detailed below.

\smallskip\noindent\textbf{Flight arena.}
Experiments are conducted within an indoor flight arena measuring (8$\times$5$\times$6)\,$\mathrm{m}^3$.
A motion capture system provides ground truth poses at \SI{100}{\hertz} via~12~cameras.

\smallskip\noindent\textbf{Hardware platform.}
The quadrotor platforms are built from off-the-shelf components and feature a Raspberry Pi 5 compute module running the open-source Agilicious framework~\cite{Foehn_2022}. This framework tracks the point mass reference trajectories generated by the high-level planners (LMPG, MPC (cvel), MPC (sopt), and MPG) via low-level geometric and INDI~\cite{sun2022comparative} controllers that account for the full quadrotor dynamics. Onboard, the drones fuse the external pose measurements with high-rate inertial data using an Extended Kalman Filter to estimate the full state. Both high-level planners and low-level controllers use this state estimate to close the loop.

\smallskip\noindent\textbf{Software architecture.}
All high-level planners and the referee system are implemented in Julia and run on an external server (MacBook Pro M1). High-level trajectories computed by the server are communicated via WebSockets to a ground station running the Agilicious safety nodes and ROS master, which then forwards the plans via Wi-Fi to the drones.

\smallskip\noindent\textbf{Results.}
Due to the spatial constraints of the flight arena, experiments are limited to the Lemniscate track and zero-shot on the Trefoil track (c.f.~\figref{fig:paper_banner_results}l). We run a total of 8 races per comparison group.
As seen in \figref{fig:paper_banner_results}c, \figref{fig:paper_banner_results}f and \figref{fig:paper_banner_results}i, the real-world flights align with the trends observed in the high-speed asynchronous simulations. 
While MPG struggles with computational delays, LMPG's accelerated approximation of game-theoretic strategies enables it to dominate the real-world tournament.
The supplementary video visualizes the overtaking dynamics and the impact of computational delays on physical hardware.

\section{Conclusion \& Future Work}
\label{sec:conclusion}

This work examines the trade-off between extensive planning to find high-fidelity strategies and expedited decision-making that ignores future interdependence of actions in the context of autonomous drone racing.
We found that, under realistic conditions, MPG outperforms MPC at moderate velocities, but loses its advantage at higher speeds due to latency.
To overcome this latency bottleneck, we introduced LMPG, an approach that shifts expensive computations to an offline training phase. 
This learning-based approach achieves inference times approximately 14~times faster than MPG and outperforms both MPG and MPC variants in extensive simulation and hardware experiments.

Future work will focus on relaxing the assumption of perfect state information by integrating onboard perception, moving towards fully autonomous racing in clutter-rich environments without external motion capture systems.
Additionally, while this work utilized a hierarchical control approach to study the benefits of accelerated decision-making within a high-level planner, future research should investigate the potential advantages of end-to-end planning and simultaneous dynamics learning, similar to approaches found in end-to-end reinforcement learning.

\bibliographystyle{ieeetr}

\bibliography{new}
\end{document}